# 基于强化学习的科学数据特征生成算法

肖濛 [1]　周骏丰 [2]　周园春 [1,2]

[1]（中国科学院计算机网络信息中心　北京　100190）

[2]（中国科学院大学　　北京　100049）

（shaow@cnic.cn）

## Reinforcement Learning-based Feature Generation Algorithm for Scientific Data

Xiao Meng[1], Zhou Junfeng[2], Zhou Yuanchun[1,2]

[1]（*Computer Network Information Center, Chinese Academy of Sciences, Beijing 100190*）

[2]（*University of Chinese Academy of Sciences, Beijing 100049, China*）

**Abstract**　Feature generation (FG) aims to enhance the prediction potential of original data by constructing high-order feature combinations and removing redundant features. It is a key preprocessing step for tabular scientific data to improve downstream machine-learning model performance. Traditional methods face the following two challenges when dealing with the feature generation of scientific data: First, the effective construction of high-order feature combinations in scientific data necessitates profound and extensive domain-specific expertise. Secondly, as the order of feature combinations increases, the search space expands exponentially, imposing prohibitive human labor consumption. Advancements in the Data-Centric Artificial Intelligence (DCAI) paradigm have opened novel avenues for automating feature generation processes. Inspired by that, this paper revisits the conventional feature generation workflow and proposes the Multi-agent Feature Generation (MAFG) framework. Specifically, in the iterative exploration stage, multi-agents will construct mathematical transformation equations collaboratively, synthesize and identify feature combinations exhibiting high information content, and leverage a reinforcement learning mechanism to evolve their strategies. Upon completing the exploration phase, MAFG integrates the large language models (LLMs) to interpretatively evaluate the generated features of each significant model performance breakthrough. Experimental results and case studies consistently demonstrate that the MAFG framework effectively automates the feature generation process and significantly enhances various downstream scientific data mining tasks.

**Keywords**　Deep Reinforcement Learning, Multi-Agent, Data-centric AI, Feature Generation, Scientific Data

**摘要**　特征生成（Feature Generation, FG）的目标是通过构建高阶特征组合并去除冗余特征，提升原始数据的预测潜力。针对表格型科学数据，特征生成是提高下游机器学习模型性能的重要预处理环节。然而，传统方法在处理科学数据特征生成问题时面临以下两方面挑战：首先，针对科学数据生成有效的高阶特征组合通常需要深入且广泛的领域知识；其次，随着特征组合阶数的增加，组合搜索空间呈指数级扩张，导致人工探索成本过高。近年来，以数据为核心的人工智能（Data-Centric Artificial Intelligence, DCAI）范式的兴起为自动化特征生成过程提供了新的可能性。受此启发，本文重新审视了传统特征生成工作流程，并提出了一种多智能体特征生成框架（Multi-agent Feature Generation, MAFG）。具体而言，在迭代探索阶段，多个智能体协作构建数学变换算式，识别并合成具有高信息含量的特征组合，最后通过强化学习机制实现策略的自适应演化。探索阶段结束后，MAFG 框架引入大语言模型（Large Language Models, LLMs），针对探索过程中的每个关键模型性能突破

收稿日期：2025 年 5 月 7 日　修回日期：2025 年 7 月 7 日

基金项目：国家自然科学基金项目（No.92470204），北京市自然科学基金项目（No.4254089）

　　　　　This work was partially supported by National Natural Science Foundation of China (No.92470204) and the Beijing Natural Science Foundation (No.4254089).

通讯作者：周园春（zyc@cnic.cn）

点，解释性评估该步骤新生成的特征。实验结果和具体案例研究表明，MAFG 框架能够有效地实现特征生成过程的自动化，并能显著提升下游多个科学数据挖掘任务的效果。

**关键词** 深度强化学习，多智能体，以数据为核心的人工智能技术，特征生成，科学数据

**中图法分类号** TP181

随着大数据的时代的到来，各个科学领域的数据集的规模不断增大，复杂性也不断上升。如何有效地利用数据集中的特征信息成为了重要研究的问题。特征生成是表格类科学数据（Tabular Scientific Data）预处理中关键的一环，例如科学家通过先验知识与数学公式，组合成高、体重等特征，构建身体质量指数（Body Mass Index，BMI）这一关键特征来诊断患糖尿病的风险。然而对于大数据时代的科学数据集而言，传统的手工特征生成方法由于对专家经验的巨大依赖，以及受限于科学数据集复杂性而导致的巨大人力成本，导致其应用范围有限。

近年来，随着人工智能技术的进步，处理各类研究任务的机器学习模型逐渐成熟，研究视角从设计更为精巧的模型转移到以数据为核心的人工智能技术，并以设计更高质量的特征[1]为目标，这为自动化的特征生成带来了新的驱动力。在这样的背景下，深度强化学习在单一智能体场景中取得了显著的成果[2]，多智能体强化学习（Multi-Agent Reinforcement Learning，MARL）作为其扩展[3,4]，逐渐被应用于多智能体协同决策问题。MARL 允许多个智能体在一个系统中独立行动，并通过与环境交互和奖励信号来优化其策略。将 MARL 应用于特征工程，可以将特征生成的决策过程建模为马尔科夫过程（Markov Decision Process，MDP），并利用多智能体的协同决策能力来利用数学变换组织特征，生成新的高效特征。各智能体可以通过与环境交互学习到有效的特征变换策略，通过奖励机制优化智能体参数，提高下游模型机器学习的性能。然而，基于强化学习范式的自动特征工程也有其天然的局限性。这主要来源于其产出的高阶变换特征难以解释，在实际生产应用中难以落地。

基于以上讨论，本研究提出了一种基于多智能体强化学习的科学数据特征生成与自我解释框架，该框架通过引入强化学习模型，将特征与特征的组合过程视为一个多智能体的 MDP 决策过程。三个智能体通过策略函数决定选择什么特征，以及通过一元或二元变换来组合、生成特征。此外，本研究引入了大语言模型，对特征生成的关键性能突破点进行了可解释性评估，提升了生成特征的科学意义。本工作的研究意义主要体现在以下几个方面，首先本设计为科学数据集的特征工程提供了一个新的视角和方法，打破了传统方法在处理数据集时的局限性。其次，本研究通过实验和案例研究，验证了所提方法的有效性和鲁棒性，

为相关科学领域的研究提供了可解释性评估。最后，该方法能够在所选数据集上实现科学数据优化，并显著提升下游机器学习任务的效果，为科学数据的挖掘和分析提供了有力支持。

# 1 相关工作

## 1.1 自动特征生成

自动特征生成（Automated Feature Generation，AFG）作为提升特征表达能力的有效手段，旨在通过数学运算、围绕数据本身，对原始特征进行空间扩展与重构，已成为机器学习预处理环节的重要研究方向[5]。当前研究主要沿着三个技术路径展开：基于扩展-缩减的方法通过显式特征生成策略[6]或贪婪搜索算法[7]对原始特征空间进行扩展，继而采用特征选择机制进行空间约简，这类方法在多项式组合生成能力和复杂特征评估效率方面存在显著局限性；进化-评估方法将特征生成与选择过程整合为闭环学习系统，借助进化算法或强化学习模型进行迭代优化[8,9,10]，但其离散决策机制导致计算复杂度高且收敛稳定性不足；而基于自动机器学习（AutoML）的范式尝试将特征工程转化为自动化学习任务[2,11]，虽然在特征构造自动化方面取得进展，但特征生成阶段受限且变换质量呈现较大波动性，并由于时间复杂度过高，难以应用于大规模数据上。此外，以上方法主要关注于生成过程，缺失了解释性评价，导致产出的高阶特征常常不具有人工可读性，优化后的数据集仅能供模型训练而无法直接应用于科学数据挖掘中最关注的知识发现。总结而言，随着特征组合阶数的提升，特征空间呈指数级扩展，将导致搜索空间和计算复杂度迅速增加。总结而言，随着特征组合阶数的提升，特征空间呈指数级扩展，导致搜索空间和计算复杂度迅速增加。尽管多智能体机制和强化学习策略在一定程度上有助于缩小探索范围，但面对极大规模或高维科学数据，组合爆炸仍可能带来巨大的计算资源消耗和搜索效率下降，影响实际应用的可扩展性。此外，现有方法对高阶变换特征的理论分析和数学有效性刻画尚不充分，缺乏系统性的解释性评价。针对这些挑战，本文引入大语言模型（Large Language Model，LLMs）作为知识驱动的解释与筛选机制，自动对生成特征进行可解释性评估，并在特征空间扩展过程中动态去除难以解释或冗余的高维特征，从而提升特征生成的科

学有效性与实际可用性。

## 1.2 深度强化学习算法

强化学习算法可以大致分为两类[1]，一种是基于模型的强化学习，另一种是无模型强化学习。对于基于模型的强化学习，顾名思义即依赖环境模型的强化学习算法。它通过利用环境模型来降低对交互数据的依赖，从而加速训练速度并提高学习效率。其典型算法有基于动态规划的 DP（Dynamic programming）算法[12]，基于模型价值迭代的 MBVI（Model-based value iteration）算法[13] 以及基于模型策略迭代的 MBPI（model-based policy iteration）算法[14]。

对于无模型强化学习，及不依赖环境模型的强化学习算法，经典算法包括 Q-Learning、SARSA（State-action-reward-state-action）等。Watkins 在 1989 年提出的 Q-Learning[15] 作为强化学习发展的基石对后面的算法起到了推动作用。在 Q-learning 的基础上，2013 年 Mnih 等人提出了深度 Q 网络 DQN（Deep Q-Learning Network）[16]，这些方法将深度神经网络引入 Q-learning 算法，并使用神经网络来估计 Q 值函数，通过这样的方式来对复杂状态空间进行学习和表示。DQN 是一种将深度学习与 Q 学习相结合的算法，它能够在高维状态空间下进行决策。其核心思想是使用神经网络来近似 Q 函数，通过经验回放和目标网络的更新来提高学习的稳定性和效率。该网络通过学习获得的奖励（reward）来对参数进行优化，以获得函数的最大 Q 值，从而使得智能体能够在动态的系统环境中学习和决策。此外还有一些将两种思想融合的算法，如 AC（actor-critic）算法、A2C（advantage actor-critic）算法、A3C（asynchronous actor-critic）算法[17,18] 等。

# 2 定义与问题陈述

## 2.1 操作语义空间

特征生成旨在通过数学变换来组合特征以生成高阶特征，设基础特征集合为 $F = \{f_1, ..., f_m\}$，特征生成的数学操作符集合 $\mathcal{O} = \mathcal{O}_1 \cup \mathcal{O}_2 \cup \mathcal{O}_3$ 由以下三类运算符构成。下面分别对这三类运算符进行详细介绍：

1）一元运算符 $\mathcal{O}_1$：对单个特征进行数学变换或预处理，$\mathcal{O}_1$ 中包含的操作有 $\sqrt{x}$，$x^2$，$\sin(x)$，$\tanh(x)$，$\text{sigmoid}(x)$，$\log(x)$，$\frac{1}{x}$。

2）二元运算符 $\mathcal{O}_2$：对两个特征进行组合运算：

$\mathcal{O}_2 = \{x + y, x - y, x \times y, x/y\}$，这些运算符可以将不同的特征进行组合，生成新的特征，从而挖掘特征之间的潜在关系。

3）预处理运算符 $\mathcal{O}_3$：预处理运算符 $\mathcal{O}_3 \subseteq \mathcal{O}_1$，包含标准化、归一化和分位数变换，这些方法是特征尺度变换操作的子集。这些运算符支持在生成过程中动态调整数据分布，使得生成的特征更适合后续的机器学习任务。其中标准化的定义为：$\text{stand\_scaler}(x) = \frac{x - \mu}{\sigma}$，这里 $\mu$ 是特征的均值，$\sigma$ 是特征的标准差。通过标准化操作，可以将特征的均值变为 0，标准差变为 1，使得不同特征具有相同的尺度，有助于机器学习模型的训练；归一化的定义为：$\text{minmax\_scaler}(x) = \frac{x - \min(x)}{\max(x) - \min(x)}$，其目标在于将特征的值域映射到[0,1]区间。归一化可以消除特征之间的量纲差异，使得所有特征在相同的尺度下进行比较。分位数变换 $\text{quan\_trans}(x)$ 通过经验分布函数映射至均匀分布，这种变换可以使得特征的分布更加平滑，减少异常值对模型的影响。

## 2.2 特征变换序列

特征变换序列是由一系列操作符和基础特征按照一定顺序组合而成的表达式。在 MAFG 框架中，特征变换序列用于描述如何从基础特征集合 $F$ 生成新的特征。例如，对于基础特征 $f_1$ 和 $f_2$，一个可能的特征变换序列可以是 $\tau = \left(\sqrt{f_1} + f_2\right)^2$。

## 2.3 状态-动作表征

在多智能体强化学习中的状态表征用于描述当前环境的信息，而动作表征用于表示智能体可以采取的行动。

状态表征需要包含足够的信息，以便智能体能够做出合理的决策。在 MAFG 框架中，状态包括当前的特征集合以及已经生成的特征变换序列。

动作表征表示智能体在当前状态下可以采取的行动。在特征生成问题中，动作是选择某一个操作符或选择操作符所相应的特征进行组合（例如一元运算符需要选择一个特征，而二元操作符则需要选择两个特征），以生成新的特征变换序列。例如，选择操作符 "+" 对特征 $f_1$ 和 $f_2$ 进行组合，就可以表示为一个动作。

通过操作符集合和操作符映射字典来实现动作的编码和解码。智能体根据当前状态并，通过相应的

策略函数，生成新的特征变换序列。

## 2.4 问题陈述

科学数据特征生成框架的目标是找到最优的特征变换序列组合，从而提高下游机器学习模型的性能。具体而言，给定原始数据集 $D = (x_i, y_i)_{i=1}^N$，其中 $x_i$ 表示样本特征，$y_i$ 表示其标签或响应变量，相关特征集合 $F$，下游机器学习模型 $M$ 及其性能评价指标 $V(\cdot)$（如准确率、AUC、MSE 等），以及一组可选的特征变换操作集合 $O$，模型目标位自动地为特征集合 $F$ 设计一条或多条特征变换序列 $\tau$。每个序列由不同的特征变换操作按照一定的顺序组成，作用于 $F$ 上生成新的特征集合 $F'$，并用于训练下游的机器学习模型，具体目标函数如下所示：

$$\tau^* = \arg\max_\tau V\left(M(D, \tau(F))\right) \quad (1)$$

## 3 深度强化学习框架

### 3.1 MAFG 整体架构设计

如图 1 所示，本研究提出的基于多智能体强化学习的科学数据特征生成框架 MAFG 通过三个协同工作的强化学习智能体实现动态特征生成与优化，其架构设计包含特征生成的探索、智能体的优化、关键特征组合的解释等模块。本节将系统性阐述各模块的设计思想与交互机制。

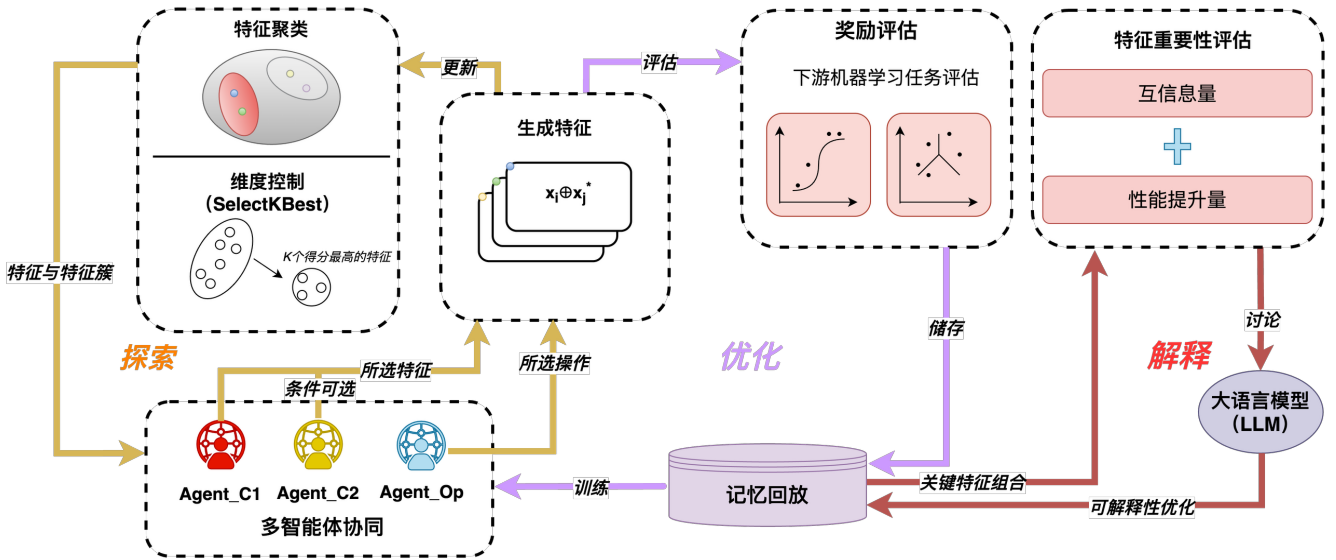

**Fig. 1 The framework of MAFG**

**图 1 MAFG 架构图**

### 3.2 数据预处理模块

本框架以来自科学数据银行（Science Data Bank）上的科学数据集为输入，并采用 MinMaxScaler 方法对数据进行归一化处理，对于输入特征矩阵 $X \in R^{n \times m}$（$n$ 为样本数，$m$ 为特征维度），其归一化公式为：

$$X_{norm}^{(i,j)} = \frac{X^{(i,j)} - \min(X_{\cdot j})}{\max(X_{\cdot j}) - \min(X_{\cdot j})} \cdot (b - a) + a$$
$$\forall i \in \{1, \ldots, n\}, j \in \{1, \ldots, m\}, \quad (2)$$

其中，$a$ 和 $b$ 分别为目标区间的下限与上限，本框架设定 $a = -1$，$b = 1$。该操作通过保留原始数据分布形态，将特征值映射至对称区间，有助于提升后续强化学习策略网络的训练稳定性。归一化后的特征矩阵 $X_{norm}$ 与目标变量 $y$ 重新组合为标准化数据集 $\mathcal{D}_{\mathscr{G}} = [X_{norm} | y]$，作为特征生成流程的输入，最终构建标准

化数据表作为后续模块的输入。

### 3.3 多智能体协同机制

一个基础的一元或者二元数学变换可以由两个必须的元素和一个候选元素组成。在此基础上，本框架设计了两个特征聚类智能体（Agent_C1, Agent_C2）与一个操作选择智能体（Agent_Op），三者通过状态-动作链式决策实现特征空间探索，以实现一元或者二元数学变换的生成，各智能体具体含义如下：

**1）Agent_C1**：基于当前特征聚类状态，从特征簇集合中选取初始特征子集。

**2）Agent_Op**：根据选定特征簇的嵌入表示，从预定义操作集 $O$ 中选择特征变换算子。

**3）Agent_C2**：在二元操作场景下动态激活，基于操作语义与初始特征簇状态选择辅助特征簇，完成

特征交叉计算。

智能体间采用参数隔离设计,通过差异化状态空间定义(如 Agent_C2 引入操作编码的扩展状态)实现决策解耦,同时共享全局奖励信号进行协同优化。

### 3.4 多智能体协同特征生成

特征生成的核心是在分层强化决策与特征簇的动态组合,在特征聚类层面,采用无监督聚类算法将高维特征空间划分为语义相近的特征簇,聚类数目通过超参数调优确定。该方法构建特征组的粗粒度表示,为强化学习提供可解释的决策单元。

在强化学习系统中,智能体的有效决策高度依赖于对环境状态的精准表征和关键特征的抽取。因此,每个智能体的输入状态是当前数据集的统计信息(例如标准差、均值、分位数等)与前步骤智能体的选择特征子集的统计量信息或所选择数学变换操作的独热编码(One-hot encoding)。

在候选组合特征的选择层面,每回合(Episode)中每一步(Step)特征选择,系统采用阶段反馈驱动的策略进行参数更新。首先由聚类智能体 Agent_C1 从特征聚类中选择一个聚类作为当前操作的特征子集,接着由操作选择智能体 Agent_Op 从操作集合中选择一个操作,如一元操作(O1)或二元操作(O2)。如果选择的是二元操作,另一个聚类智能体 Agent_C2 会从特征聚类中选择另一个聚类作为第二个操作数。当生成特征维度超过预设阈值时,使用基于互信息统计量的特征选择算法进行特征重要性排序,保留对目标变量预测贡献最高的特征子集,保留最重要的特征,有效控制特征空间爆炸问题。三个智能体基于 DQN 系列算法[19,20]。DQN 是一种将深度学习与 Q 学习相结合的算法,它能够在高维状态空间下进行决策。其核心思想是使用神经网络来近似 Q 值函数,通过经验回放和目标网络的更新来提高学习的稳定性和效率。

在 DQN 中,使用一个神经网络 $Q(s, a; \theta)$ 来估计状态 $s$ 下执行动作 $a$ 的 Q 值,其中 $\theta$ 是神经网络络的参数。训练的目标是最小化 Q 值估计与目标 Q 值之间的均方误差,即:

$$L(\theta) = E_{(s,a,r,s')\sim D}\left[\left(r + \gamma \max a' \ Q\left(s', a'; \theta^-\right) - Q(s, a; \theta)\right)^2\right], \quad (3)$$

其中,$D$ 是经验回放缓冲区,$\theta^-$ 是目标网络的参数,$\gamma$ 是折扣因子。在下文中,我们还将介绍基于 DQN

算法的改进算法,作为智能体强化学习方法上的策略优化方向。

最后,在特征组合层面,依据选定操作对特征簇实施符号计算,例如对两个特征簇进行逐元素相加生成高阶特征。该过程引入特征命名规则化机制,保障特征的可追溯性。在组合特征之后,如果特征的总数超过了阈值,则会对特征集合进行维度控制,采用特征选择算法智能地选择最优的特征子集。

### 3.5 动态评估与优化机制

强化学习奖励机制旨在通过下游预测任务的性能提升量,构建即时评估函数,将特征生成性能提升效果量化为强化学习信号。对于分类类型的下游任务(Classification Task),使用随机森林分类器,采用分层 K 折交叉验证,计算加权 F1 分数的平均值:

$$P_{new}^{cls} = \frac{1}{5}\sum_{k=1}^{5} F1_k^{weighted}, \quad (4)$$

$$F1_k = 2 \cdot \frac{precision_k \cdot recall_k}{precision_k + recall_k}, \quad (5)$$

式中,$F1_k^{weighted}$ 表示第 $k$ 折交叉验证中按类别样本量加权的 F1 分数,计算覆盖全部类别。

对于回归类型的下游任务(Regression Task),框架使用随机森林回归器,计算相对绝对误差(1-RAE)的平均值,性能指标如下:

$$P_{new}^{reg} = \frac{1}{5}\sum_{k=1}^{5}(1 - RAE_k), \quad (6)$$

$$RAE_k = \frac{\sum_{i\in D_k}|y_i - \hat{y}_i|}{\sum_{i\in D_k}|y_i - \overline{y_{train}}|}, \quad (7)$$

$D_k$ 为第 $k$ 折测试集,$\overline{y_{train}}$ 为训练集目标变量均值,$RAE_k$ 反映预测值 $\hat{y}_i$ 相对于均值基准的相对绝对误差。

对于异常检测类型的下游任务(Anomaly Detection),框架将使用 K 近邻分类器,计算 ROC AUC 分数的平均值,性能指标如下:

$$P_{new}^{det} = \frac{1}{5}\sum_{k=1}^{5} AUC_k, \quad (8)$$

$$AUC_k = \int_0^1 TPR_k(FPR) \, dFPR, \quad (9)$$

$TPR_k(FPR)$ 为第 $k$ 折的 ROC 曲线,$AUC_k$ 表示模型区分正负类的能力。

对于以上三种任务,最终奖励函数可统一表达为:

$$R_t = \eta \cdot (P_{new} - P_{old}), \quad (10)$$

其中，$P_{new}$ 与 $P_{old}$ 按任务类型选择公式(4)(6)(8)计算，$\eta \in [0,1]$ 为任务自适应的奖励缩放因子（默认 $\eta = 1$）。计算得到的奖励信号会通过分配给不同的动作或状态信号，并存储到相应的深度 Q 网络（DQN）的经验回放缓冲区中，用于后续的学习和更新。

各智能体会独立地维护经验回放缓冲区，存储状态转移元组（状态、动作、奖励、下次状态）。当缓冲区达到容量阈值后，采用批量采样方式更新 Q 网络参数，通过目标网络延迟更新策略提升训练稳定性。

### 3.6 基于大语言模型的特征评估模块

针对所生成的高阶特征评估的科学性和解释性问题，本文进一步设计并集成了基于大语言模型的特征评估模块。如图 1 右侧所示，MAFG 系统在每一轮训练或决策阶段，会从 memory（记忆库）中动态抽取最近的状态-动作-奖励历史序列，并结合奖励信息和决策动作，对相关特征进行评估。

为了定量指导特征生成过程、避免因智能体拟合产生过于高阶组合特征的"幻觉"问题，系统将关键特征及其变换结果输入大语言模型，由模型输出并关于特征重要性、相关性和作用机制的多维评价。具体而言：1）对于每一组特征，框架将奖励分数自动量化其对下游任务表现的影响力，避免仅凭大语言模型输出文本进行主观解读；2）通过与已有领域知识的交叉验证，特征评估模块对特征的合理性与科学性进行知识佐证，显著减少因框架拟合生成幻觉解释的风险。3）框架将会去除过于复杂或是高奖励分但无法被大语言模型解释的特征变换，进一步提高特征评价的可靠性和可解释性。

综上，基于大语言模型的特征评估模块在强化学习系统中实现了对特征价值的定量判别和风险防控，有效避免了 MAFG 框架解释幻觉问题，并为后续特征筛选、特征变换和决策优化提供了科学支持，提升了系统的泛化能力与整体解释性。

### 3.7 小结

综上综述，MAFG 框架的特征分组策略主要包含以下几个步骤：首先对特征进行聚类，将相似的特征划分到同一个组中。然后使用强化学习模型从聚类结果中选择合适的特征组。最后根据选择的操作对特征组进行处理，生成新的特征。在产生关键的特征变换时，外置的大语言模型将会对这些关键变换进行可解释评价，从而使得变换过程中产生的知识能够被有效利用。通过框架自动探索特征空间，选择最有价值的特征组进行特征生成，从而提高下游机器学习模型的性能。

## 4 实验结果与性能评估

### 4.1 实验数据集选择

本设计经过多维度调研，最终在科学数据银行[1]（Science Data Bank）上挑选出三个实验对象数据集：气象与环境因子交互驱动的手足口病发病率数据集[21]、学习投入-师生关系协同驱动的学业表现数据集[22]、肾功能指标交互作用驱动的衰弱指数数据集[23]。这些数据集均为包含连续性回归预测任务的表格类型数据集（tabular data），且分别对应了公共卫生领域、教育领域和临床医学领域。

### 4.2 性能分析与评估

在模型训练的过程中，日志会记录下每一节（Episode）中每一步（step）的性能表现（performance），并与之前的最优表现进行比较并更新最优表现。在性能指标方面，由于本研究选取的回归评价指标主要有以下三个：MAE（Mean Absolute Error）、RMSE（Root Mean Squared Error）、1-RAE（1-Relative Absolute Error）

平均绝对误差（MAE），也称为 L1 损失，是损失函数的一种。其计算方式是，先求出预测值与实际值之间的绝对差值，再对整个数据集的这些差值取平均值。MAE 衡量误差的大小，不考虑误差的方向。一般来说，MAE 越低，模型的准确性就越高。RMSE 也称为均方根偏差，RMSE 是通过取 MSE 的平方根来计算的。RMSE 越低，模型预测效果就越好。相对绝对误差（1-RAE）取的是总绝对误差除以平均值和实际值之间的绝对差的值。RAE 以比率的形式呈现，误差值在 0 到 1 之间，并且越接近 1 越好。

在手足口病气象数据集的训练日志中，Episode 1-Step 3 生成特征 temperature × precipitation 使模型 1-RAE 从基准值 0.302 跃升至 0.358，提升幅度达 18.5%（其余两项指标提升不大），是全局性能最高跃升点。

表 1 手足口病气象数据集性能分析表

| 性能指标 | 原始数据 | MAFG | 变化幅度 |
|---|---|---|---|
| 1-RAE | 0.302 | 0.358 | +18.5% |
| MAE | 37.385 | 37.366 | -0.05% |
| RMSE | 5983.257 | 5713.384 | -4.51% |

在小学生学习投入数据集的训练日志中，Episode 2-Step 4 生成特征 ZT3dedication[3] × ZT2TS[2]）/ ZT3absorption 使模型 RMSE 从 0.013 降至 0.004 下降幅度为 69.2%。该特征在验证集的 MAE 从 0.079

---

降至 0.038，误差下降幅度为 51.8%。

**Table 2 Performance analysis table of primary school learning input data set**

**表 2　小学生学习投入数据集性能分析表**

| 性能指标 | 原始数据 | MAFG | 变化幅度 |
|---|---|---|---|
| 1-RAE | 0.896 | 0.953 | +6.3% |
| MAE | 0.079 | 0.038 | -96.2% |
| RMSE | 0.013 | 0.004 | -51.8% |

在胱抑素 C 与衰弱风险数据集的训练日志中，Episode 4-Step 4 生成特征 (newcrp² + newhba1c) × cystatinc 使模型 1-RAE 从基准值 0.02 跃升至 0.035，提升幅度达 16.5 倍，是全局性能最高跃升点。性能对比具体如下表：

**Table 3 Performance analysis table of Cystatin C and frailty risk data set**

**表 3　胱抑素 C 与衰弱风险数据集性能分析表**

| 性能指标 | 原始数据 | MAFG | 变化幅度 |
|---|---|---|---|
| 1-RAE | 0.002 | 0.035 | +1650% |
| MAE | 8.500 | 8.188 | -3.6% |
| RMSE | 122.857 | 119.254 | -2.9% |

总结而言，本实验通过对比 MAFG 框架生成特征的下游机器学习表现能力，与原始数据的下游任务表现能力论证了框架架构的有效性，在后续实验中，将会探讨强化学习算法的选择、框架中特征冗余去除模块，以及具体的性能跃升点的案例分析。

### 4.3 不同强化学习优化算法对比实验

本实验过改变框架智能体的强化学习方法做横向对比试验，并在三组不同类型的数据集上，通过相同轮次的探索与学习，对比了 DQN、DDQN、DuelingDQN 和 DuelingDDQN 四种强化学习方法在特征评价任务中的表现，并以原始数据的性能作为基线进行分析。结果如图 2 所示。首先可以发现各方法在不同数据集上的 1-RAE（相对绝对误差的倒数）均显著高于原始数据，表明基于强化学习的特征评价方法能够有效提升模型性能。经实验验证模型性能如下图：

由图可见不同强化学习算法在不同数据集上的优化效果不尽相同。具体而言，在手足口病气象数据集上，DQN 方法的 1-RAE 为 0.358，优于原始数据的 0.302，DDQN 和 DuelingDQN 的性能略有提升，分别达到 0.373 和 0.375，DuelingDDQN 取得了最高的 1-RAE（0.392），提升最为显著。在小学生学习投入数据集中，四种方法的表现进一步提升。DQN 达到 0.917，DDQN 和 DuelingDQN 均为 0.936，DuelingDDQN 最高，为 0.950，均高于原始数据的 0.896，显示出强化学习方法对于复杂行为类数据的良好适应性。在胱抑素 C 与衰弱风险数据集上，虽然整体 1-RAE 数值较小，但各方法同样优于原始数据（0.002）。DQN、DDQN 和 DuelingDQN 分别为 0.009、0.010 和 0.010，DuelingDDQN 提升至 0.012，显示出强化学习方法在医学指标风险评估场景下的有效性。

总体而言，DuelingDDQN 在相同轮次的探索和学习过程中，在所有数据集上均取得了最佳性能，验证了其在特征抽取与评价方面的优势。同时，基于各种强化学习方法的 MAFG 框架均能显著提升模型的表现，进一步证明了本框架的有效性和通用性。

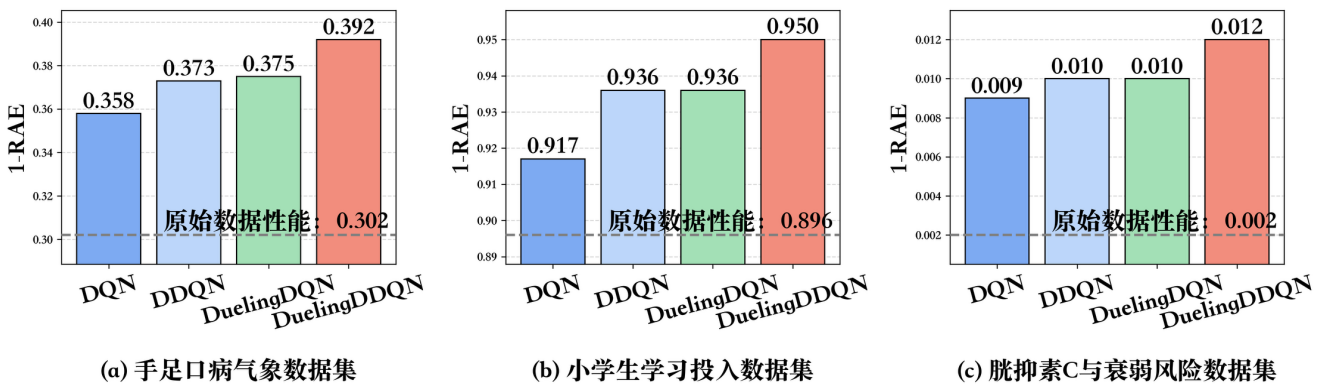

(a) 手足口病气象数据集　　　(b) 小学生学习投入数据集　　　(c) 胱抑素C与衰弱风险数据集

**Fig. 2 Comparison of performance changes under different DQN algorithms**

**图 2　不同 DQN 算法数性能变化对比图**

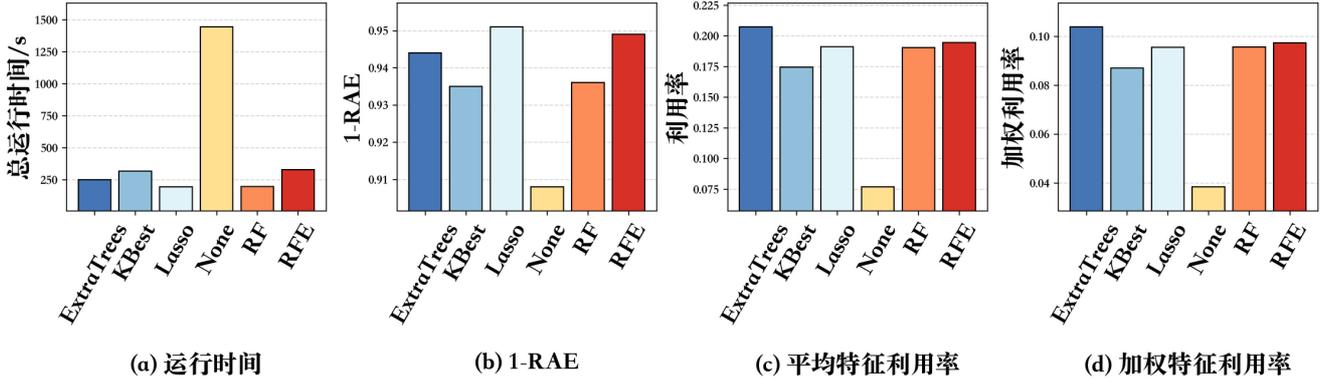

(a) 运行时间　　　(b) 1-RAE　　　(c) 平均特征利用率　　　(d) 加权特征利用率

**Fig. 3 Comparison of different feature selection component**

**图 3 不同特征选择组件的对比图**

## 4.4 特征选择组件对比实验

本实验以学习投入-师生关系协同驱动的学业表现数据集为实验对象,旨在系统评估所提出的多智能体特征生成(MAFG)框架在表格型教育数据中的特征冗余去除能力和高价值特征挖掘能力。实验重点考察 MAFG 生成的新特征在不同特征选择方法辅助下对提升下游学业表现预测模型的实际效果与特征利用率的贡献,从而验证数据为中心的特征生成方法在科学数据建模中的实用价值。

在本实验中,MAFG 分别结合 ExtraTrees、KBest、Lasso、RandomForest(RF)、Recursive Feature Elimination(RFE)以及 None(不使用特征选择)六种特征选择方式,构建对比实验组。此外,为全面衡量新生成特征的实际利用率,设计了两类特征利用率评价指标:

**平均特征利用率(Average Proportion)**:衡量每轮 top-k 特征中,新生成特征(即不属于原始特征集)的占比,定义为:

$$\text{Proportion} = \frac{|\{f \in T | f \notin F\}|}{|T| + \epsilon} \quad (11)$$

其中 $T$ 为当前轮 top-k 特征集合,$F$ 为原始特征集合,$f$ 为一个特征,$\epsilon$ 为极小常数。越大的特征利用率 Proportion 值表示本轮所生成的特征在 Top-K 重要特征中的占比越大,代表着生成特征更有效。

**加权特征利用率(Reward-weighted Proportion)**:在平均特征利用率基础上,引入每轮性能提升信号,突出高性能场景下新特征的贡献,定义为:

$$\text{Weighted Proportion} = \text{Proportion} \times \sigma(r) \quad (12)$$

其中 $r$ 为当前轮 reward,$\sigma(r) = \frac{1}{1+e^{-r}}$ 为 Sigmoid 函数。越大的加权特征利用率表示本轮生成的特征不但能提升模型性能,同时也更为高效。

每项指标均取整个实验周期的平均值,以反映

各特征选择方法下 MAFG 框架的特征冗余去除和高价值特征生成能力。

图 3 中实验结果表明,MAFG 框架结合各类特征选择器均能提升下游 1-RAE 性能,并在合理的运行时间内完成特征生成。尤其是在 ExtraTrees、Lasso 等方法下,特征生成的效果最为显著。相比之下,None 组(不使用特征选择模块)的运行时间明显更长,表明特征选择模块的去冗余功能能够有效加速特征生成过程。对比 None 组方法与其他特征选择组件的 1-RAE 分数,也充分展示了特征选择组件能够通过去除低质量的生成特征,提升最终生成数据集的品质。在特征利用率方面,平均特征利用率与加权特征利用率均显示,MAFG 生成的新特征在大多数特征选择器辅助下更容易被下游模型采纳,且高性能场景下新特征实际贡献进一步提升。这表明所提出方法不仅能够去除冗余特征,还能有效挖掘和利用高价值特征。

总结而言,本实验系统验证了多智能体特征生成框架中特征选择组件的价值。综合对比表明,MAFG 框架联合特征选择策略,不仅能够提升模型预测性能,还能显著提升新特征的利用率和实际价值。这一结果有力支持了数据为中心的特征生成理念在实际应用场景中的有效性和普适性。

## 4.5 案例分析

本小节将围绕三组科学数据集进行结果讨论,结合模型定量的性能评估与特征评估模块,分析生成的特征组合在具体案例下的实际意义。

### 4.5.1 气象与环境因子交互驱动的手足口病发病率预测优化

用 K-Best 特征选择算法得出的数据集原特征重要性如下表:

**Table 4 Importance table of features of hand-foot-mouth dis-ease meteorological data set**

**表 4　手足口病气象数据集原特征重要性表**

| 特征名称 | 重要性 |
|---|---|
| population | 0.180 |
| temperature | 0.286 |
| humidity | 0.105 |
| air pressure | 0.069 |
| precipitation | 0.219 |
| wind speed | 0.054 |
| sunshine hour | 0.202 |

注：“population”表示报告病例所在地区的人口总数；“temperature”表示病例所在地区在报告病例当天的气温；“humidity”表示病例所在地区在报告病例当天的相对湿度；“air pressure”表示病例所在地区在报告病例当天的气压；“precipitation”表示病例所在地区在报告病例当天的降水量；“wind speed”表示病例所在地区在报告病例当天的风速。

经模型训练得到的训练日志反映出性能随步数增长所发生的变化趋势与关键性能变化步的特征生成组合，如下图所示：

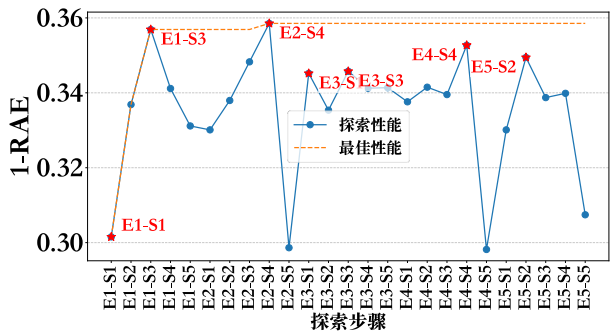

**Fig. 4 Performance variation map of the hand-foot-mouth disease dataset**

**图 4　手足口病气象数据集性能变化图**

其中标注出的关键性能变化步的特征生成组合如下表：

**Table 5 Table of generated feature combinations for hand-foot-mouth disease meteorological dataset**

**表 5　手足口病气象数据集生成特征组合表**

| 特征组合 | 步骤 | 重要性 |
|---|---|---|
| cube(population) | E1,S1 | 0.199 |
| stand_scaler(air_pressure) | E1,S3 | 0.101 |
| temperature × precipitation | E2,S4 | 0.281 |
| reciprocal(sunshine_hour) | E3,S1 | 0.067 |
| sigmoid(air_pressure+population) | E3,S3 | 0.207 |

| precipitation×air_pressure | E4,S4 | 0.218 |
| tanh(temperature) | E5,S2 | 0.017 |

综合重要性和性能表现，选取重要性排名第一的特征组合 temperature×precipitation，结合特征评估模块对该特征组合进行讨论：

**Table 6 The environmental significance of generated feature combination**

**表 6　特征组合环境意义讨论结果**

| 特征 | 环境意义 | 文献支持 |
|---|---|---|
| temper-ature | 温度升高加速病毒复制与媒介生物活动 | 《PLoS NTD》指出温度>25°C 时 EV71 病毒活性提升 3 倍（doi:10.1371/journal.pntd.0003692）[24] |
| precipi-tation | 降水增加形成积水环境，促进蚊媒滋生 | 《Environ Health Perspect》证实降水与手足口病传播呈 J 型曲线关系（doi:10.1289/EHP13807）[25] |
| temper-ature×precipi-tation | 高温高湿环境产生协同放大效应 | 《Sci Total Environ》发现温雨乘积项比单因素预测效能高 41%（doi:10.1016/j.scitotenv.2022.160850）[26] |

### 4.5.2　学习投入-师生关系协同驱动的学业表现预测优化

用特征选择模块得出的数据集原特征重要性如下表：

**Table 7 Original feature importance table of the primary school students' learning engagement dataset**

**表 7　小学生学习投入数据集原特征重要性表**

| 特征名称 | 重要性 |
|---|---|
| gender | 0.0086 |
| T3age | 0.0075 |
| T3grade | 0.0094 |
| ZT2PS | 0.1784 |
| ZT2TS | 0.1663 |
| ZT2PR | 0.1051 |
| ZT3vigor | 0.1966 |
| ZT3dedication | 0.1853 |
| ZT3absorption | 0.1830 |

注：“gender”表示性别；“T3age”指第三次数据收集时的年龄；“T3grade”指第三次数据收集时的年级；“ZT2PS”　指标准化后的积极亲子风格得分；“ZT2TS”　指标准化后的积极师生关系得分；“ZT2PR”　指标准化后的

同伴关系得分；"ZT3vigor"指"活力"维度得分；"ZT3dedication"指"奉献"维度得分；"ZT3absorption"指"专注"维度得分。

经模型训练得到的训练日志反映出性能随步数增长所发生的变化趋势，如下图：

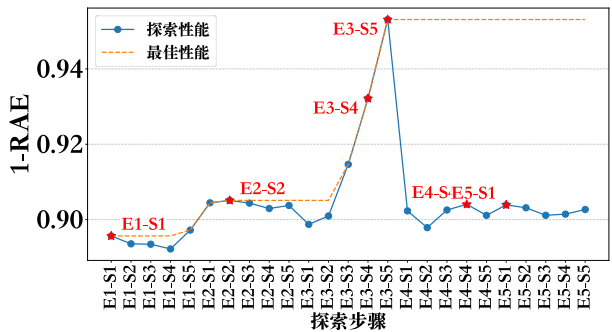

**Fig. 5 Performance variation of primary school students' learning engagement data set**

**图 5 小学生学习投入数据集性能变化图**

其中标注出的关键性能变化步的特征生成组合如下表：

**Table 8 Table of generated feature combinations for the primary school students' learning engagement dataset**

**表 8 小学生学习投入数据集生成特征组合表**

| 特征组合 | 步骤 | 重要性 |
|---|---|---|
| ZT2PS ÷ gender | E1,S1 | 0.0686 |
| ZT2TS × ZT2PR | E2,S2 | 0.1346 |
| sqrt(ZT3absorption × ZT3dedication) | E3,S4 | 0.1929 |
| (ZT3dedication³ × ZT2TS²) / ZT3absorption | E3,S5 | 0.1742 |
| sqrt(ZT3vigor − ZT3absorption) | E4,S4 | 0.0932 |
| Sigmoid(ZT2PS + T3age) | E5,S1 | 0.1270 |

综合重要性和性能表现，选取重要性排名第二的特征组合(ZT3dedication³ × ZT2TS²) / ZT3absorption，结合特征评估模块对该特征组合进行讨论：

**Table 9 Discussion of the generated features from the primary school students' learning engagement dataset**

**表 9 小学生学习投入数据集生成特征意义讨论结果**

| 特征 | 教育学意义 | 文献支持 |
|---|---|---|
| ZT3dedication³ | 立方放大"奉献精神"的阈值效应，>4.2分时产生"心流体验"持续效应 | 《Learning and Instruction》证实奉献度³比线性项更能预测学业成就(doi:10.1016/j.learninstruc.2024.102054)[27] |
| ZT2TS² | 二次项捕捉师生关系的边际递增效应，优质师生互动对学业的促进作用 | 《Educational Psychology》发现师生关系的预测效能比线性高29%(doi:10.1080/01443410. |

呈加速趋势

| | | |
|---|---|---|
| ZT3absorption | 分母设计体现"过度专注可能抑制师生互动获益"的调节效应 | 2024.2387544)[28]《Frontiers in Psychology》报道专注度>7分时需平衡师生互动(doi:10.3389/fpsyg.2021.771272)[29] |

从讨论结果来看，生成特征通过量化"奉献³×师生²/专注"的协同机制，突破传统模型的解释瓶颈，为建立"以学定教"的精准教育干预体系提供量化工具。

### 4.5.3 肾功能指标交互作用驱动的衰弱指数预测优化

用特征选择模块得出的数据原特征重要性如下表：

**Table 10 Original feature importance table of the Cystatin C and fateful risk dataset**

**表 10 胱抑素 C 与衰弱风险数据集原特征重要性表**

| 特征名称 | 重要性 |
|---|---|
| cystatinc | 0.0299 |
| newcrp | 0.0038 |
| newhba1c | 0.0010 |
| newglu | 0.0094 |
| newcho | 0.0049 |
| newbun | 0.0075 |
| newcrea | 0.0058 |

注："ID"为识别序号，"cystatinc"表示胱抑素 C，一种肾功能指标；"newbun"表示血液中的尿素氮；"newglu"表示血液中的葡萄糖；"newcrea"表示血液中的肌酐；"newcho"表示血液中的胆固醇等指标；"newcrp"表示 C 反应蛋白，一种炎症标志物；"newhba1c"表示糖化血红蛋白，用于评估糖尿病控制情况；"newua"表示尿分析结果。

经模型训练得到的训练日志反映出性能随步数增长所发生的变化趋势，如下图：

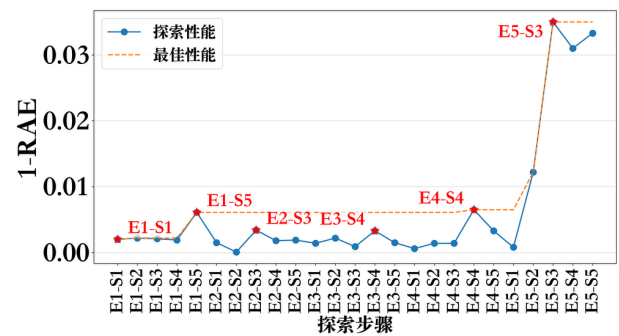

**Fig. 6 Cystatin C and frailty risk data set performance change map**

**图 6 胱抑素 C 与衰弱风险数据集性能变化图**

其中标注出的关键性能变化步的特征生成组合如下表：

**Table 11 Table of the generated feature combinations for Cystatin C and frailty risk datasets**

**表 11　胱抑素 C 与衰弱风险数据集生成特征组合表**

| 特征组合 | 步骤 | 重要性 |
|---|---|---|
| sqrt(cystatinc) | E1,S1 | 0.0078 |
| cystatinc³ | E1,S5 | 0.0056 |
| cystatinc × newcrp | E2,S3 | 0.0132 |
| minmax_scaler(cystatinc) | E3,S4 | 0.0025 |
| sigmoid(cystatinc) | E4,S4 | 0.0034 |
| (newcrp² + newhba1c) × cystatinc | E5,S3 | 0.0093 |

综合重要性和性能表现，选取重要性排名第二的特征组合(newcrp² + newhba1c) × cystatinc，结合特征评估模块对该特征组合进行讨论：

**Table 12 The biological significance of the generated feature combinations**

**表 12　特征组合生物学意义讨论结果**

| 特征 | 教育学意义 | 文献支持 |
|---|---|---|
| new-crp² | CRP 的平方项反映炎症的指数级放大效应 | 《Kidney International》指出 CRP² 可预测 CKD 患者心血管事件风险（doi: 10.1016/j.kint.2018.11.045）[30] |
| newhba1c | 长期血糖控制指标，高值导致晚期糖基化终产物(AGEs)堆积，加速肾间质纤维化 | 《Diabetes Care》证实 HbA1c 与肾小球硬化面积正相关（doi:10.2337/diacare.23.4.544）[31] |
| cysta-tinc | 肾小球滤过率敏感指标，立方操作可能对应肾单位损伤的临界点效应 | 《Nephrology Dialysis Transplantation》提出 cystatinc³ 可识别早期肾功能非线性衰减（doi:10.1093/ndt/gfae06932）[32] |

特征评估模块揭示了 (炎症² + 代谢紊乱) × 肾功能损伤的系统效应，捕捉到 CKD 患者衰弱进展的核心病理三角，炎症与代谢紊乱的协同作用被肾功能衰减放大，形成恶性循环。显著提升模型预测能力，并为医学临床的潜在精准干预提供新靶点。

## 5 结论

本文针对大数据时代科学数据集特征复杂、人工特征生成方法局限性突出的现状，结合以数据为核心的人工智能视角，提出了一种基于多智能体强化学习与大语言模型相结合的自动特征生成与自我解释框架。通过将特征生成过程建模为多智能体马尔科夫决策过程，各智能体能够协同探索和优化特征组合策略，显著提升了特征工程的自动化和有效性。引入大语言模型对关键特征及其变换过程进行评估与解释，不仅增强了模型的可解释性，也提升了生成特征的科学合理性和实用价值。实验结果表明，所提方法在手足口病气象、小学生学习投入、胱抑素 C 与衰弱风险等多个科学数据集上均显著优于原始数据集，验证了本方法在特征生成和下游任务建模方面的有效性与鲁棒性。此外，特征评估模块为特征的重要性分析和科学性验证提供了有力支持，推动了科学数据智能处理的进一步发展。

未来研究可进一步拓展多智能体强化学习在异构数据、多模态场景下的特征生成能力，并探索大语言模型与领域知识更加深度的融合，为科学数据挖掘和知识发现提供更为强大的智能工具。

**作者贡献声明：**肖濛提出了整体算法框架，进行了论文初稿撰写；周骏丰完成了实验代码，参与了论文初稿撰写；周园春进行了论文终稿修改。

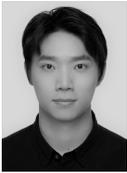
**Xiao Meng,** born in 1995. PhD, Assistant Researcher, Member of CCF. His main research interests include AI4Science and Data-centric AI.

肖濛，1995 年生。博士，助理研究员，CCF 会员。主要研究方向为科学智能，以数据为核心的人工智能技术。

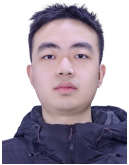
**Zhou Junfeng,** born in 2002. Undergraduate student. His main research interests include data mining.

周骏丰，2002 年生。本科生。主要研究方向为数据挖掘。

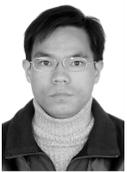
**Zhou Yuanchun,** born in 1975. PhD, professor, PhD supervisor, Senior Member of CCF. His main research interests including bigdata and scientific data mining.

周园春，1975 年生。博士，研究员，博士生导师，CCF 高级会员。主要研究方向为科学数据与数据智能。